\documentclass[9.5pt,journal,final,finalsubmission,twocolumn]{IEEEtran}
\usepackage{epsf,amsmath,amssymb,graphicx,epsfig,subfigure}
\usepackage{hyperref}
\usepackage{csquotes}
\usepackage[lined, algonl]{algorithm2e}
\usepackage{amsmath}
\usepackage{booktabs}

\title{Information-Theoretic Aggregation of Ethical Attributes in Simulated-Command}

\author{
    \IEEEauthorblockN{Taylan Akay\IEEEauthorrefmark{1}, Harrison Tolley\IEEEauthorrefmark{1}, Hussein Abbass\IEEEauthorrefmark{1}}\\
    \IEEEauthorblockA{\IEEEauthorrefmark{1} School of Systems and Computing, University of New South Wales, Canberra ACT, Australia.\\
    Emails: \{t.akay, h.tolley, h.abbass\}@unsw.edu.au}
}

\begin{document}

\graphicspath{ {./images/} }

\maketitle
\setcounter{page}{1}
\begin{abstract}
Ethics has been a purely human endeavour. As currently the most intelligent species with the capacity to perform the roles and duties of moral agents, humans have been carrying the responsibility to make ethical judgement. In the age of artificial intelligence (AI), human commanders need to use the computational powers available in today\textquoteright s environment to simulate a very large number of scenarios. Within each scenario, situations occur where different decision design options could have ethical consequences. Making these decisions reliant on human judgement is both counter-productive to the aim of exploring very large number of scenarios in a timely manner and infeasible when considering the workload needed to involve humans in each of these choices.

In this paper, we move human judgement outside the simulation decision cycle. The human input becomes more paramount pre-simulation, during the design of the ethical metric space used to assess generated options during the simulation, and post-simulation, during the selection of the final course of action for command. Basically, the human will design the ethical metric space, leaving it to the simulated environment to explore the space. When the simulation completes its testing cycles, the testing environment will come back to the human commander with a few options to select from. The human commander will then exercise human-judgement to select the most appropriate course of action, which will then get executed accordingly.

We assume that the problem of designing metrics that are sufficiently granular to assess the ethical implications of decisions is solved. Subsequently, the fundamental problem we look at in this paper is how to weight ethical decisions during the running of these simulations; that is, how to dynamically weight the ethical attributes when agents are faced with decision options with ethical implications during generative simulations. The multi-criteria decision making literature has started to look at nearby problems, where the concept of entropy has been used to determine the weights during aggregation. We draw from that literature different approaches to automatically calculate the weights for ethical attributes during simulation-based testing and evaluation. We then compare the resultant weights and conclude the paper.
\end{abstract}

\section{Introduction}

\par{Competent commanders understand that ethics shape their decisions. The competency of a commander lies not only in their ability to comprehend complex situations and see through risks and consequences, but also lies in their virtue of being humane, applying proportional responses, and maintaining world order by respecting the law including the Law of Armed Conflict (LOAC), sometimes referred to as the International Humanitarian Law (IHL). When it comes to ethical decisions, humans can weight their consequences and carry the burden of accountability. Hard ethical decisions are those with tensions between mission objectives and what is right. When the two are in conflict, or when options exist where some ethical principles are obeyed while others are not, trade-offs need to be made.}

\par{Artificial Intelligence (AI) promises, among others, speed of command. AI calls for rigorous testing of methods and approaches. The space of possibilities explored by AI systems is normally huge, beyond human abilities to systematically scrutinise each possibility to assess operational consequences, let alone ethical ones. To test these AI systems, simulation-based testing offers an approach where scenarios get simulated, the AI sub-systems get assessed for their performance, consequences get evaluated, and the testing procedure stress-tests the system by generating even more complex scenarios. Different philosophies, like massive scenario generation~\cite{davis2007enhancing}, and methodologies, such as computational red teaming (CRT)~\cite{abbass2015computational, tang2019simulation}, for this approach have been proposed in the literature.
	

\par{The challenge in using any of these automated testing approaches is how to assess a solution during simulation when the assessment involves ethical attributes. In theory, it is possible we could design metrics that act as proxies for ethical attributes. For example, if an ethical principle is to minimise harm to non-combatants, we could possibly categorise harm into physical, functional, economic, psychological, and social. We could then design one or more measures for each category of harm. For some ethical principles and situations, this process could become cumbersome, possibly without a satisfactory solution. Consider for example the nuisances that exist in quantifying mission effects and the trade-offs that need to be made with the estimation of casualties to estimate proportionality. While we acknowledge the complexities associated with the design of ethical attributes and indicators, we assume in this paper that an acceptable solution is found. For those principles where we can\textquoteright t design these metrics, we will reserve them for the commander to assess before choosing the final course of action. However, there remains a fundamental problem to solve, which is the focus of this paper: how to combine those ethical metrics we managed to acceptably quantify within our automated test and evaluation (T\&E) environment to guide the test-cases\textquoteright  generation process?}

\par{Luckily, the multi-criteria decision making (MCDM) and multi-attribute decision analysis (MADA) literature have investigated this problem for decades, but with the assumption that there is a human who can provide input to this weighting process. A Priori method expects the human decision maker to reveal their preferences a priori; thus, if these preferences are represented by weights, these methods assume that the weights to combine the attributes are known before solutions get generated. A posteriori method will first generate a set of solutions, such as a set of non-dominated solutions, then invite the human decision maker to choose from the set. Interactive methods generate a dialogue, whereby cycles of computer-generated solutions and human-judgement get repeated until a decisive solution is chosen. When there are too many attributes, it is infeasible to use a posteriori method because non-dominated and efficient solutions get very sparse. A mechanism to weight the attributes while unifying their dimensionality and magnitudes reduces the problem to a single objective, or at best to a problem with a very few number of objectives. While a linear weighted sum of objectives could face theoretical problems depending on concavity of the Pareto frontier, these theoretical challenges have been overcome by relying on different non-linear weighted-fusion methods.}

\par{The MCDM and MADA literature have assumed that there is a human to generate the weights for the fusion process. Attempts have been made to take the human out of this loop by relying on theoretical properties of the \enquote{wanted} or \enquote{desired} solutions, see for example the Technique for Order of Preference by Similarity (TOPSIS)~\cite{tzeng1981multiple}. Nevertheless, the problem of how to determine the weights to combine the attributes during CRT and other automated and simulated T\&E methods remain largely an open area of research. This is the problem we will be addressing in this paper.}

\par{The remainder of this paper is organised as follows. A short review of the multicriteria decision making literature is presented in Section~\ref{sec:back}. Different methods to automatically calculate the weights during T\&E are then proposed in Section~\ref{sec:method}, followed by a use-case to illustrate the calculations in Section~\ref{sec:usecase}. Conclusions are drawn, and future work are discussed, in Section~\ref{sec:conc}.}

\section{Multicriteria Decision Making}\label{sec:back}

\par{Decision theory relies on three fundamental premises: decision-makers invariably have a deep comprehension of their issues; these issues can consistently be framed as matters of effectiveness and the essential information and resources for finding a solution are unceasingly available. In practice, none of these hypotheses holds true as noted by Bouyssou et al.~\cite{bouyssou2006evaluation}. Decision-makers often lack an accurate comprehension of their issues; frequently, these issues can be articulated as the pursuit of a compromise, and the resolution of a problem is invariably limited by the constraints of available resources.}

\par{Multi-criteria decision-making methods (MCDM) require a structured organisation of issues tied to decisions by pinpointing options and the characteristics that distinguish them. This systematic approach facilitates the elucidation of the problem and enhances the efficacy of alternative ranking. Human insight is highly valuable in MCDM frameworks for navigating decision dilemmas, specifying criteria, and extracting preferences. A variety of human insight forms has been vital to MCDM techniques such as the Analytic Hierarchy Process (AHP)~\cite{saaty1980analytical}, TOPSIS~\cite{tzeng1981multiple}, Elimination and Choice Translating Reality (ELECTRE)~\cite{benayoun1966electre}, the Preference Ranking Organisation Method for Enrichment Evaluation (PROMETHEE)~\cite{brans1982elaboration}, and Multi-Attribute Utility Theory (MAUT)~\cite{fischer1972four}, among others. Recent advancements have led to innovative MCDM methodologies that rectify the shortcomings of conventional approaches. Multi-objective optimisation problems form a critical subset of MCDM techniques, where conflicting objectives can be formulated and solved to generate a non-dominated option set. Frameworks like the Weighted Sum Model (WSM), the Weighted Product Model (WPM), and Fuzzy TOPSIS contribute to selecting from the Pareto-optimal set~\cite{Sahoo2023}. 

\par{The judgment aggregation frameworks employed in these methodologies depend on expert knowledge to prioritise criteria and arrive at a collective decision~\cite{anniciello2023judgment}. For instance, AHP allows decision-makers to establish the priority of various criteria through pairwise comparisons. Human decision-makers articulate their preferences regarding criteria and through eigenvalue calculations~\cite{podvezko2007determining}, human decision makers can consolidate their evaluations into a singular score for each alternative.}

\par{The development of the MAUT stemmed from applying a multi-linear and quasi-additive utility framework. This approach was originally formulated to incorporate multiple objectives, intangible aspects, risk considerations, qualitative data, and temporal influences during ex-ante assessments, all aligned with the preferences of human decision-makers~\cite{bossel1976multiattribute}. MAUT employs utility functions to systematically quantify the preferences of decision-makers across various criteria. The fundamental concept is to allocate a utility value to each alternative. These utilities are constructed as approximations of human judgement, while encapsulating overall preference or value considering all criteria. These utility values could yield a singular synthesised score for each alternative, thereby facilitating comparative analyses~\cite{taherdoost2023multi}.}

\par{When considering MAUT, the emphasis on maximising utility, with individual attributes linked to a utility value, and the comprehensive utility for a choice is obtained by aggregating these separate values. This framework indicates that the preferences of the human decision-maker making the decisions can be embedded through a utility function that integrates these qualities. The utility associated with each attribute is conventionally normalised within the range of 0 to 1 (i.e. normalised utility function), with 0 denoting the outcome with least utility and 1 signifying the outcome with the highest utility~\cite{von1973multi}. In a fashion comparable to various MCDM strategies, MAUT involves determining weights for each attribute to illustrate their significance. Weights may be established through the AHP or by directly soliciting preferences from human decision-makers~\cite{papageorgiou2016value}. The weights serve as scaling constants that indicate the relative contribution of each criterion\textquoteright s utility to the overall problem space, and their derivation from a foundational utility function guarantees that they are anchored in the value trade-offs of the decision-maker as opposed to being assigned arbitrarily. The overall utility of an alternative is computed by aggregating the products of the attribute utilities and their respective weights.}

\par{Ascertaining the suitable weights for various attributes is of paramount importance, as it significantly affects the final decision. The existing body of literature presents a multitude of instances wherein methodologies for weight derivation are extensively classified into expert judgment~\cite{pena2020explicit}, pairwise comparisons such as the AHP~\cite{odu2019weighting}, entropy-based approaches~\cite{liu2021determining}, discriminating power techniques~\cite{fu2018determining}, and hybrid methodologies that amalgamate these approaches~\cite{yin2016combination}.}

\section{Methodology}\label{sec:method}

\subsection{Problem Definition}

\par{A utilitarian approach requires external judgement on the importance of each criterion in a decision-making context. In the absence of human input to guide automated T\&E, information theory could provide a basis for judgement by analysing the behaviour of attributes across T\&E scenarios. The core idea is that an attribute\textquoteright s discriminatory power is reflected in higher-order statistics of its behaviour. For example, if we assess the monetary cost of some alternative choices and the variance of the cost is very small, the cost becomes a less important attribute to consider. Similarly, if casualties remain high across all military scenarios, resulting in low variance, casualties contribute less to the decision and should be given less weight. Furthermore, we need ways to contrast the different variances arising from evaluating different attributes in the simulation environment. This suggests that information theoretic measures could offer a solution.}

\par{MAUT has a vector of indexed alternatives ($m = 1, \dots,M$), and each alternative has a vector of indexed attributes ($r = 1, \dots, R$). An alternative in our context is a course of action. Examples of these courses of actions that may impose ethical dilemmas include whether to allow a military vehicle to follow a path on a road with children-crossing, disengage a weapon from firing at a high-value target next to a highly populated area with civilians, or vacating an injured soldier in a time-critical situation that does not allow proper assessment of precautionary measures.

In case of simulation runs for automated T\&E, the attributes either aggregate to, or are, the measures of performance generated during different simulation runs. Without loss of generality, we will not differentiate between attributes and measures of performance in the remainder of this paper. A vector of indexed attributes ($r = 1, \dots, R$) get assessed over a vector of indexed situations ($s = 1, \dots, S$). A measurement function ($E$) returns the value of an attribute in a situation; that is, $E: (r,s) \rightarrow R$. Without loss of generality, we will assume that each measurement function can be transformed to, or it is in itself, a normalised utility function, such that, there is a utility assessment for each attribute-situation pair: $U: (r,s) \rightarrow [0,1]$.

For each attribute, $r$, we will denote the first non-central moment (i.e. mean) and second-central moment (i.e. variance) over the utility space as $\dot{U}(m,r)$ and $\ddot{U}(m,r)$, respectively. These quantities follow the classic formulaes for untabulated data as below:

\[ \dot{U}(m,r) = \frac{\sum_{s=1}^{S} U(m,r,s)}{S} \]

\[ \ddot{U}(m,r) = \frac{\sum_{s=1}^{S} (U(m,r,s) - \dot{U}(m,r))^2}{S-1} \]

We can now define two matrices. The first matrix, $\Gamma$, is a matrix of expected utilities, while the second matrix, $\Lambda$, is the second-central moments of these utilities.

\[
\Gamma =
\begin{bmatrix}
    \dot{U}(1,1)        & \ldots    & \dot{U}(1,R) \\
    \vdots              & \ddots    & \vdots \\
    \dot{U}(M,1)        & \ldots    & \dot{U}(M,R) \\
\end{bmatrix}
\]

\[
\Lambda =
\begin{bmatrix}
    \ddot{U}(1,1)        & \ldots    & \ddot{U}(1,R) \\
    \vdots              & \ddots    & \vdots \\
    \ddot{U}(M,1)        & \ldots    & \ddot{U}(M,R) \\
\end{bmatrix}
\]

The question we are attempting to answer in this paper is how to assign weights for $\Gamma$ using information gleaned from the simulation runs; in other words, how can we assign weights that carry meaningful information on the importance of an attribute? Our premise is that the second-central moment is a source of information that could be used for this purpose; that is, weights need to be derived from $\Lambda$, whereby each attribute is weighted based on the nature of variations it exhibits on different courses of actions. An attribute should not be weighted high if it has similar variance on all courses of actions. For example, consider an attribute such as \enquote{risk to the troops} (RTT). If the variance of RTT in simple ethical situations is the same as in complex ethical situations, such an attribute should be weighted less when making a decision. However, if the variance of RTT is high in complex ethical situations and low in simple ethical situations, RTT carries more information and has a stronger discriminatory power in this case than the previous case of similar variances.

In the remainder of this section, we propose three methods to transform the second non-central moment information into weights for different attributes. In the following section, we will present a scenario to illustrate each method numerically, while presenting in the appendix a step-by-step calculations of the inner working of each method.

\subsection{Contribution}

\par{Classically, the utilisation of entropy weighting (EW) concerns an assignment of weights independent of subjective assessments. EW reflects variabilities in the data. As pointed out by Chen~\cite{chen2021effects}, the resultant weights do not necessarily reflect importance of attributes per se. If subjective importance are different, then EW needs to be combined with subjective importance. EW quantifies the level of uncertainty or randomness intrinsic to the data where attributes exhibiting greater variability are assigned lower weights, signifying their diminished relevance in the decision-making framework~\cite{arshad2024decision}. This approach proved to be especially advantageous in contexts characterised by substantial data variability or in instances where subjective evaluations are either unavailable or lack reliability~\cite{pliego2024integrated}.}

\par{Our proposed approach differs from the above on multiple grounds. Classic approaches for EW assume that there is a crisp measure for each attribute on each alternative. Variabilities are then calculated on the level of these crisp measures; in other words, the variabilities relates to the changes in how each alternative performs for each attribute. In the notations we used in the previous section, this is equivalent to variabilities in the first non-central moment, $\Gamma$.}

\par{In simulation-based T\&E, there are two sources of variabilities for an attribute. The first arises from the variability of the performance of the attribute in different situations and scenarios relative to a particular alternative; this is denoted above by $\Lambda$. The second is variabilities on alternatives, or $\Gamma$. As Chen~\cite{chen2021effects} suggested, the second source of variability does not indeed reflect importance of an attribute. Variabilities in $\Gamma$ discount the weights in classic EW methods, where an assumption that an attribute should not be sensitive to changes in alternatives. While this could be challenged in its own right, attributes need to have discriminatory powers; we use them to assess and monitor a phenomenon about a problem and therefore, they should discriminate different situations and contexts. This point holds true for the first source of variability, where a high variability for an attribute-alternative pair across different situations and scenarios entails richer information contents for that attribute.}

\par{The novelty of our approach stems from the fact that we separate the information that we use to derive the weights ($\Lambda$) from the information we use to make decisions ($\Gamma$). Basically, the weights represent the information contents provided by an attribute based on measurements and/or utilities drawn from assessing it in different situations and contexts, but for the same alternative. The discriminatory power of an attribute represents a measure of importance.}

One could still argue that while our EW calculations present importance information, the nature of these importance information is still different from subjective importance. This is true. However, due to our aim to combine ethical attributes, we assume that the ethical measurements have been approved by the decision maker. This approval process should entail two factors. First, that the measure is an acceptable proxy to the phenomenon it measures. Second, that the range, and therefore the magnitude, of the measure reflects the importance of this ethical dimension. Our reliance on utilities, $U(r,s)$, and not on measurements, $E(r,s)$, assumes that the decision maker\textquoteright s subjective judgement has been incorporated in the construction of these utilities.}

\subsection{Weight Estimation Methods}

The previous sub-section suggested that the weights of an attribute need to reflect the information contents of that attribute in its ability to discriminate among different alternatives. Fortunately, information theory offers a large number of methods to assess these information contents. We will focus this paper on a few low-cost estimates and use a hypothetical use case to demonstrate the weights they estimate.

First, we will need to calculate $p(m,r)$ that is the relative variance of alternative $m$ for attribute $r$ as follows:

\[ p(m,r) = \frac{\ddot{U}(m,r)}{\sum_{i=1}^{M} \ddot{U}(i,r)} \]

\subsubsection{ICW: Information Contents Weights}

    \[E(r) = \sum_{i=1}^M - p(i,r) \log_2 p(i,r) \]

    The weight of the $r^{th}$ attribute as delineated in the matrix representation above can be computed as:

    \[ICW(r) = \frac{E(r)}{\sum_{j=1}^R E(j)}, \ \ r=1, \ldots, R\]

\subsubsection{IGW: Information Gain Weights}

    The Kullback-Leibler divergence (KLD), also known as information gain, is a measure of the distance between two distributions. The literature adopted variations of KLD. For example, Quinlan~\cite{quinlan1993c} adopted KLD for splitting on continuous variables by measuring the actual difference between the entropy before and after a split. This suggests that we could adopt KLD in multiple forms. We propose two forms: one relies on a prior distribution drawn from the subjective belief of the decision maker and the other relies on a prior distribution drawn from the performance of all attributes.

    In the first form, we call it IGH to emphasise its reliance on human input, one distribution will be drawn from the variance matrix $\Lambda$ similar to ICW. The second distribution is a pre-defined subjective assessment. Basically IGH will combine the discriminatory power of an attribute as measured within a simulation environment with the subjective belief of its importance as measured by the decision maker. The further away the simulation-driven probabilities are from the subjective ones, the higher the weight.

    \[ IGH(r)= D(p(i,r) \,\|\, q(i,r))   = \sum_{i=1}^M  p(i,r) \log \frac{p(i,r)}{q(i,r)}  \]

    The weight of the $r^{th}$ attribute as delineated in the matrix representation above can be computed as:

    \[IGHW(r) = \frac{IGH(r)}{\sum_{j=1}^R IGH(j)}, \ \ r=1, \ldots, R\]

    In the second form, we call it IGD to emphasise its reliance on data without human involvements, we calculate IGD as the difference between the entropy of an attribute and the average entropy of all other attributes.

    \[ IGD = E(r) - \sum_{j=1, j \ne r}^R (1-E(j)) \]

    The weight of the $r^{th}$ attribute as delineated in the matrix representation above can be computed as:

    \[IGDW(r) = \frac{IGD(r)}{\sum_{j=1}^R IGD(j)}, \ \ r=1, \ldots, R\]

\subsection{Approach Adoption for Autonomy}\label{sec:extension}

While the approach is presented for automated T\&E purposes, it is important to mention that it is not limited to this use alone. Simulations are used by autonomous systems as a way to think about the world. An autonomous system may need to run many scenarios in their computational brain before making a decision. The autonomous system would then run through scenarios with different force protection and proportionality levels, autonomously derive the weights then make a decision such as whether to fire at a target or not. The proposed approach could still be used within an autonomous system to automatically derive weights for attributes. The approach is flexible to account for a prior defined by a human subject, but can equally work without such a prior.

\section{Use Case}\label{sec:usecase}

\subsection{Scenario}

\par{We adopt a synthetic use case aimed at appraising the ethical decision-making of Autonomous Weapon Systems (AWS). A light enemy (RedForce) armored formation is slowly approaching the position of a friendly force (BlueForce) during a patrol mission. The BlueForce AWS is operating under Rules of Engagement (RoE) that permit action only if the AWS can positively distinguish targets from non-targets and ensure high confidence in classification metrics. No known civilians or non-combatants are detected in the area by either onboard sensors or external intelligence feeds. The following RoE are considered to be part of BlueForce\textquoteright s AWS ethical decision space aligned:
	\begin{itemize}
		\item Target detection: The EO/IR sensors identify armored vehicles moving in formation, matching hostile force profiles with 85\% target precision and 75\% recall, based on current classifier performance.
		\item Non-target detection: Terrain scans and image signatures detect no indications of civilian presence, yielding 95\% non-target precision and 92\% recall for non-combatant detection.
		\item Accuracy: The minimum deployable overall classifier accuracy for this engagement scenario is 88\%, based on validation against known target datasets (this is assumed for the concept; it is a metric that is known a priori from the model development phase).
	\end{itemize}}

\par{The AWS must analyse the battlefield environment by extracting key observables that effectively contribute to the ethical decision space. The AWS has a trusted target recognition and tracking sub-systems, capable of detecting, localising, classifying, and tracking the target and assessing the military advantage from engaging with that target; basically, the system can provide reliable scores for the force protection and proportionality attributes. The AWS may face situations that could be characterised by two attributes: Force Protection and Proportionality when applying force. The former approximates the military advantage to engage; the higher the utility of force protection, the higher the probability the AWS is expected to engage. The latter approximates the ethical decision space; the higher the utility for proportionality, the more likely the AWS is permitted to engage. These abstractions are designed for simplicity to avoid overwhelming the academic work with operational details that do not contribute to the main objective of this paper. The key decision consideration, however, is the obvious tradeoff that needs to be made in certain environments or scenarios, and the tension between operational objectives and ethical obligations in some of the resultant scenarios.}

\par{The BlueForce AWS needs to be assessed on two different operational scenarios (Scenario 1 and Scenario 2). The first scenario involves varieties of situations where force protection is more prominent but also ethical considerations sit at a high stake. The second scenario involves more ethically complex situations where the AWS is less likely to engage. To test and evaluate (T\&E) the AWS on both scenarios, a digital twin of the AWS will run within a generative simulation evaluation environment capable of dynamic generation of diverse situations in both scenarios. The AWS will be exposed to a large number of situations sampled from each scenario space. The outcome of the automated T\&E environment should be a ranking of the two scenarios in terms of their expected utilities; the higher the expected utility, the more favourable it is to use the AWS in that scenario.}

\subsection{Results}

\par{In each simulation run, the AWS is exposed to a situation with different combinations of force protection and proportionality. The system makes decisions on whether it should strike or not. The utility associated with force protection and proportionality are recorded for each decision and for each simulation run. 1200 samples were drawn from a distribution as an approximation of the simulation environment, with 300 samples for each alternative-attribute pair. The average and variance of these values are reported in Table~\ref{tab1}. Both the data driven probabilities (normalised variances) and subjective probabilities provided as a prior by a subject matter expert are listed in Table~\ref{tab2}. The variances were normalised to derive the simulation-driven probabilities, which are presented along side the subjective probabilities in Table~\ref{tab2}.

\begin{table}
\caption{Average and variance (between parenthesis) of utilities for each alternative-attribute pair.}\label{tab1}
\begin{center}
\begin{tabular}{lcc}
\hline
\textbf{Alternatives} 	& \textbf{Force Protection}	& \textbf{Proportionality}\\ \hline
AI1	                    &    15 (7.0)                    &  4  (1.0)                         \\
AI2                     &    8 (2.5)                    &  10 (4.0)                         \\ \hline
\end{tabular}
\end{center}
\end{table}

\begin{table}
\caption{Data driven and human (between parenthesis) subjective probabilities for each alternative-attribute pair.}\label{tab2}
\begin{center}
\begin{tabular}{lcc}
\hline
\textbf{Alternatives} 	& \textbf{Force Protection}	& \textbf{Proportionality}\\ \hline
AI1                     &    0.74 (0.7)                 &  0.2 (0.1)                        \\
AI2                     &    0.26 (0.3)                 &  0.8 (0.9)                        \\ \hline
\end{tabular}
\end{center}
\end{table}

As demonstrated in Table~\ref{tab2}, the subjective probabilities seem to match the data driven ones for force protection, while there is a clear difference for the proportionality attribute.

The AI system with maximum expected utility is the most preferred. The expected utility for each alternative is calculated as below:

\[  \mathtt{Scenario 1} = w_1 * 15 + w_2 * 4 \]

\[  \mathtt{Scenario 2} = w_1 * 8 + w_2 * 10 \]

Undertaking the calculations for each type of weight methods, Table~\ref{tab3} lists the resultant weights and the corresponding ranking for each decision.

\begin{table}
\caption{The resultant weights according to each of the three proposed approaches and the ranks resultant from the adoption of each weighting method.}\label{tab3}
\begin{center}
\begin{tabular}{lcc}
\hline
\textbf{Method} 	& \textbf{Force Protection}	& \textbf{Proportionality}\\ \hline
IC-Weight           &   0.54    &   0.46 \\
IGH-Weight          &   0.07    &   0.93 \\
IGD-Weight          &   0.50    &   0.50 \\ \hline
\textbf{Method} 	& \textbf{Scenario 1}	& \textbf{Scenario 2}\\ \hline
ICW-Expectation     &   9.94    &   8.92 \\
IGHW-Expectation    &   4.77    &   9.86 \\
IGDW-Expectation    &   9.5     &   9.0 \\ \hline
ICW-Ranking         &   1       &   2 \\
IGHW-Ranking        &   2       &   1 \\
IGDW-Ranking        &   1       &   2 \\ \hline
\end{tabular}
\end{center}
\end{table}

The results reveal the importance for incorporating human judgement into the decision making process. If the human expectation is likely to be significantly different from the analysis of variations, human judgement could get incorporated into IGH as a priori, and before running the simulations; thus, without the need for human input in each simulation rule. The human subjective probabilities could be elicited before the simulation-based T\&E exercise then the data-driven probabilities estimated during the comparisons get interpreted in light of the subjective probabilities. This could influence the chosen alternative during the simulation runs.

\section{Conclusion}\label{sec:conc}

In automated test and evaluation, a large number of situations and scenarios get assessed within a simulation environment. Ethical attributes are good examples of intangibles that need to be quantified to guide the search algorithm towards areas where high risk situations could occur. In this paper, we tackled the problem of how to automatically derive the weights to aggregate measurements of these ethical attributes. We proposed the use of information theoretic measures derived from the variance of the performance of each attribute in diverse situations and scenarios. One of the proposed methods incorporates a prior from human judgement; thus allowing the calibration of a human prior with the outcomes of the simulations. Our future work will focus on the measurement system to provide indicators for ethical dimensions such as proportionality.

\section{Acknowledgement}

This work is partially supported via the U.S. Department of Defense (DoD) and under sub-contract to University of New South Wales (UNSW). Any opinions, findings, and conclusions or recommendations expressed in this paper are those of the authors and do not reflect the view of the DoD, the U.S Government, or UNSW.

\bibliography{references}
\bibliographystyle{IEEEtran}

\appendices

\renewcommand{\thetable}{\Alph{section}\arabic{table}} 
\setcounter{table}{0}

\section{Step-by-Step Calculations for Information-Theoretic Weight Derivations}

This appendix presents detailed calculations for deriving various information-theoretic metrics based on two scenarios (Scenario 1 and Scenario 2).

\begin{itemize}
\item \textbf{Dataset Overview}

The datasets consist of two attributes for each scenario: Force Protection and Proportionality.

\item \textbf{Mean and Variance Calculations}

The following table summarises the mean and variance for Force Protection and Proportionality in each scenario:

\begin{table}[htbp]
	\centering
	\caption{Mean and Variance per Scenario and Attribute}
	\label{tab:mean_variance}
	\begin{tabular}{@{}lcc|cc@{}}
		\toprule
		& \multicolumn{2}{c|}{Scenario 1} & \multicolumn{2}{c}{Scenario 2} \\
		\cmidrule(lr){2-3} \cmidrule(lr){4-5}
		Attribute & Mean & Variance & Mean & Variance \\
		\midrule
		Force Protection & 15 & 7.0 & 8  & 2.5 \\
		Proportionality  &  4 & 1.0 & 10 & 4.0 \\
		\bottomrule
	\end{tabular}
\end{table}

\item \textbf{Normalised Variance and Logarithmic Transformation}

The variance values are normalised across scenarios for each attribute:

\begin{table}[htbp]
	\centering
	\caption{Normalised Variance and Natural Logarithm}
	\label{tab:normalised_variance}
	\begin{tabular}{@{}lcc|cc@{}}
		\toprule
		& \multicolumn{2}{c|}{Normalised Variance} & \multicolumn{2}{c}{Ln (Normalised)} \\
		\cmidrule(lr){2-3} \cmidrule(lr){4-5}
		Attribute & Scenario 1 & Scenario 2 & Scenario 1 & Scenario 2 \\
		\midrule
		Force Protection & 0.7368 & 0.2632 & -0.3054 & -1.3350 \\
		Proportionality  & 0.2    & 0.8    & -1.6094 & -0.2231 \\
		\bottomrule
	\end{tabular}
\end{table}

\item \textbf{Entropy and ICW (Information Contents Weight)}

To derive the ICW weights, the entropy of each attribute was first computed using its normalised variance across scenarios. This reflects how much uncertainty or information each attribute contributes. The entropies were then normalised, producing the ICW weights.

\begin{table}[htbp]
	\centering
	\caption{Entropy and Information Contents Weight (ICW)}
	\label{tab:entropy_icw}
	\begin{tabular}{@{}lcc@{}}
		\toprule
		Attribute & Entropy & ICW \\
		\midrule
		Force Protection & 0.5763 & 0.5353 \\
		Proportionality  & 0.5004 & 0.4647 \\
		\bottomrule
	\end{tabular}
\end{table}

\item \textbf{IGH and IGHW (Information Gain Weights)}

IGH weights were derived by comparing the simulation-driven probabilities to a set of subjective prior probabilities provided by a human subject matter expert. The more an attribute's simulated behaviour diverged from prior expectations, the higher its IGH score. These raw IGH values were then normalised to produce IGHW.

\begin{table}[htbp]
	\centering
	\caption{Information Gain (IGH) and Normalised Weight (IGHW)}
	\label{tab:igh_ighw}
	\begin{tabular}{@{}lcc@{}}
		\toprule
		Attribute & IGH & IGHW \\
		\midrule
		Force Protection & 0.0014 & 0.0694 \\
		Proportionality  & 0.0193 & 0.9306 \\
		\bottomrule
	\end{tabular}
\end{table}

\item \textbf{IGD and IGDW (Information Gain Difference Weights)}

IGD was derived by comparing the entropy of each attribute to the average informativeness of the others. This captures how uniquely informative each attribute is within the context of the full attribute set. The IGD scores were then normalised to produce IGDW.

\begin{table}[htbp]
	\centering
	\caption{Information Gain Difference (IGD) and Normalised Weight (IGDW)}
	\label{tab:igd_igdw}
	\begin{tabular}{@{}lcc@{}}
		\toprule
		Attribute & IGD & IGDW \\
		\midrule
		Force Protection & 0.0767 & 0.5000 \\
		Proportionality  & 0.0767 & 0.5000 \\
		\bottomrule
	\end{tabular}
\end{table}

\item \textbf{Expectations and Ranking}

Each weighting scheme was used to derive an expectation score for each scenario by applying the weights to the mean utility values of the attributes in that scenario. These expectations serve as composite scores, allowing each scenario to be ranked according to the specific weighting approach.

\textbf{ICW-Expectation:} The weighted sum of attribute means in each scenario using the ICW weights.

\textbf{IGHW-Expectation:} The same, but using IGHW weights derived from deviations from human priors.

\textbf{IGDW-Expectation:} Averages using the IGDW weights, which reflect relative informativeness.

The scenario with the highest expectation score under each method was given Rank 1. This ranking highlights how sensitive the decision preference is to the chosen method of weighting.

\end{itemize}

\end{document}